\documentclass{esannV2}
\usepackage{graphicx}
\usepackage{amssymb,amsmath,array}
\usepackage{subcaption}

\usepackage{lmodern}
\usepackage{hyperref}

\DeclareMathOperator{\relu}{\mathrm{ReLU}}
\DeclareMathOperator{\signsplit}{\mathrm{Split}}
\DeclareMathOperator{\smoothsplit}{\mathrm{SmoothSplit}}
\DeclareMathOperator{\parametricsplit}{\mathrm{ParametricSplit}}
\DeclareMathOperator{\prelu}{\mathrm{PReLU}}

\usepackage{tabularx}   
\usepackage{multirow}   
\usepackage{booktabs}   

\usepackage{makecell}   
\usepackage{array}
\begin{document}
\title{Topology-Aware Activation Functions in Neural Networks}

\author{Pavel Snopov$^1$ and Oleg R. Musin$^1$
  %
  %
  \vspace{.3cm}\\
  %
  University of Texas Rio Grande Valley - School of Mathematical and Statistical Sciences \\
  Brownsville - USA
}

\maketitle

\begin{abstract}
  This study explores novel activation functions that enhance the ability of neural networks to manipulate data topology during training. Building on the limitations of traditional activation functions like \( \relu \), we propose \( \smoothsplit \) and \( \parametricsplit \), which introduce topology <<cutting>> capabilities. These functions enable networks to transform complex data manifolds effectively, improving performance in scenarios with low-dimensional layers. Through experiments on synthetic and real-world datasets, we demonstrate that \( \parametricsplit \) outperforms traditional activations in low-dimensional settings while maintaining competitive performance in higher-dimensional ones. Our findings highlight the potential of topology-aware activation functions in advancing neural network architectures. The code is available via \url{https://github.com/Snopoff/Topology-Aware-Activations}.
\end{abstract}

\section{Introduction and Related Work}

Despite significant advances, the underlying mechanisms of neural network learning remain only partially understood.
Studies such as \cite{top_of_DNN} have shown that a well-trained network (i.e., one achieving near-zero generalization error) progressively transforms a complex dataset \(M = M_a \cup M_b\) into a topologically simpler one.
This transformation is largely attributed to the non-injective nature of \(\relu\), which tends to <<glue>> points together, whereas functions like \(\tanh\) preserve the input topology.

Additional work \cite{why_DL_works, sep_and_geo, activation_lands, magai2023deep} suggests that untangling latent manifolds is crucial for improving classification performance, emphasizing the role of topological transformations in deep learning.
Moreover, topology-aware activation functions have been applied successfully in tasks such as image segmentation \cite{baxter:hal-03001770} and graph neural networks \cite{9231732}.

Motivated by these insights, we propose novel non-homeomorphic activation functions that <<split>> data manifolds---effectively <<cutting>> topology rather than simply compressing it. The functions, \( \smoothsplit \) and \( \parametricsplit \), are designed to enhance a network’s ability to restructure its internal representations in a topology-aware manner.
Through experiments on both synthetic and real-world datasets, we demonstrate that \( \parametricsplit \) notably improves performance in low-dimensional settings while remaining competitive in higher dimensions, underscoring the potential of \textit{topology-aware activation functions} for advancing neural network architectures.

\section{Non-Homeomorphic Activation Functions}

As discussed in the introduction, the effectiveness of \(\relu\) (as detailed in \cite{top_of_DNN}) is largely due to its topological properties when considered as a function from \(\mathbb{R}\) to \(\mathbb{R}\). Since \(\relu\) is not a homeomorphism, it alters the topology by compressing it; specifically, it <<eliminates>> non-trivial cycles in the homology of the underlying data manifold, thereby simplifying its structure.

Topology simplification, however, can also be achieved by <<cutting>> the data manifold. When applied appropriately (e.g., along non-trivial cycles), this process divides the manifold into simpler components. A function capable of such cutting must also be non-homeomorphic; unlike \(\relu\), which is non-injective, it must be non-surjective. For example, consider the function, termed \(\signsplit\):
\[
  \signsplit(x) = x + \mathrm{sign}(x)c,
\]
where \(c\) is a learnable parameter, initialized randomly between 0 and 1, that controls the distortion at \(x = 0\). When used as an activation function, \(\signsplit\) divides the original data manifold along each dimension.

Although effective for splitting data, \(\signsplit\) is non-differentiable and therefore unsuitable for use in neural networks. To overcome this limitation, we propose a smooth approximation, \(\smoothsplit\), defined as
\[
  \smoothsplit(x) = x + \tanh(\alpha x)c,
\]
where \(\alpha\) (optimized during training and initialized randomly between 0 and 1) determines the sharpness of the approximation. For sufficiently large \(\alpha\), \(\smoothsplit\) closely approximates \(\signsplit\) while maintaining differentiability, thus making it compatible with gradient-based training.

While splitting data manifolds can simplify their topology, neural networks may still require the ability to <<glue>> points together, particularly in the final layers. To enable both splitting and compressing operations, we introduce a parametric activation function, \(\parametricsplit\), defined as:
\[
  \parametricsplit(x) =
  \begin{cases}
    bx + b\cos a - \sin a, & \text{if } x \le -\cos a,            \\[1mm]
    x\tan a,               & \text{if } -\cos a \le x \le \cos a, \\[1mm]
    x + \sin a - \cos a,   & \text{if } x \ge \cos a.
  \end{cases}
\]
This function can emulate \(\relu\), \(\signsplit\), or \(\smoothsplit\) under appropriate parameter settings:
\begin{itemize}
  \item For \(a=0,\, b=0\), \(\parametricsplit(x)\) approximates \(\relu(x-1)\).
  \item For \(a=\frac{\pi}{2},\, b=1\), \(\parametricsplit(x)\) recovers \(\signsplit(x)\).
  \item For \(a\in\left[\frac{\pi}{4}, \frac{\pi}{2}\right)\) with \(b=1\), \(\parametricsplit(x)\) approximates \(\smoothsplit(x)\), with \(\alpha\) uniquely defined by \(a\).
\end{itemize}

\begin{figure*}[!htbp]
  \centering
  \captionsetup[subfigure]{justification=centering}
  \begin{subfigure}[b]{0.24\textwidth}
    \centering
    \includegraphics[width=\linewidth]{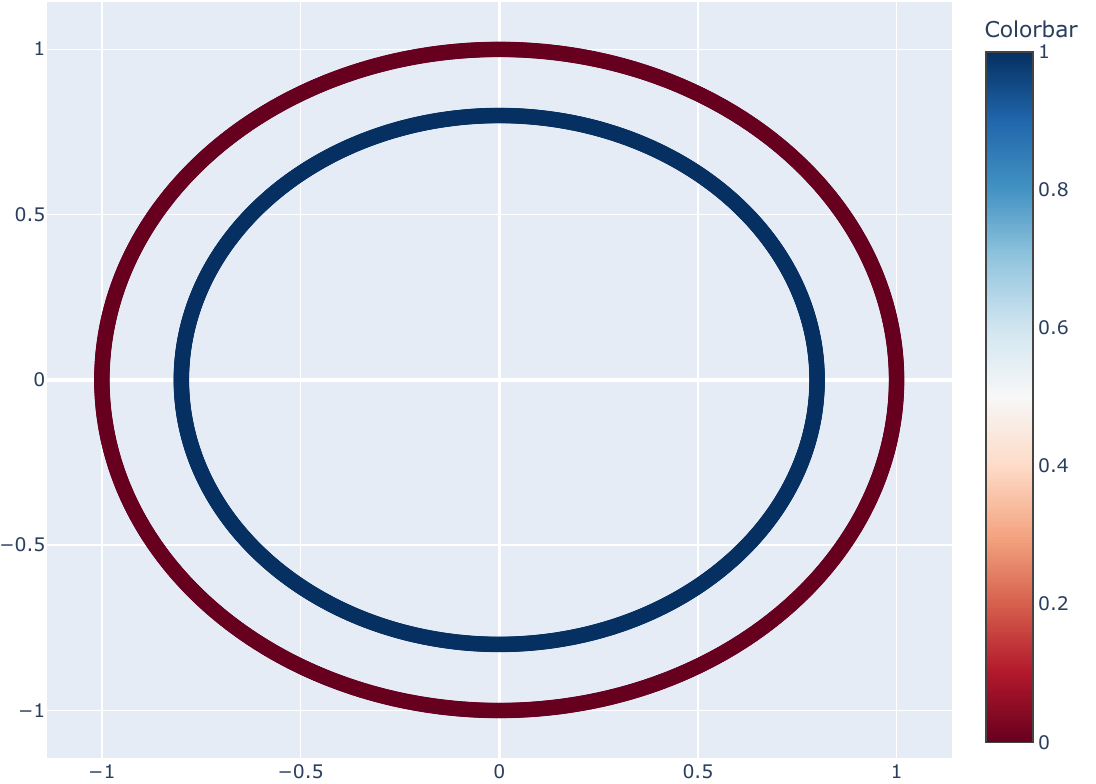}
    \caption{Original Circles dataset}
  \end{subfigure}
  \begin{subfigure}[b]{0.24\textwidth}
    \centering
    \includegraphics[width=\linewidth]{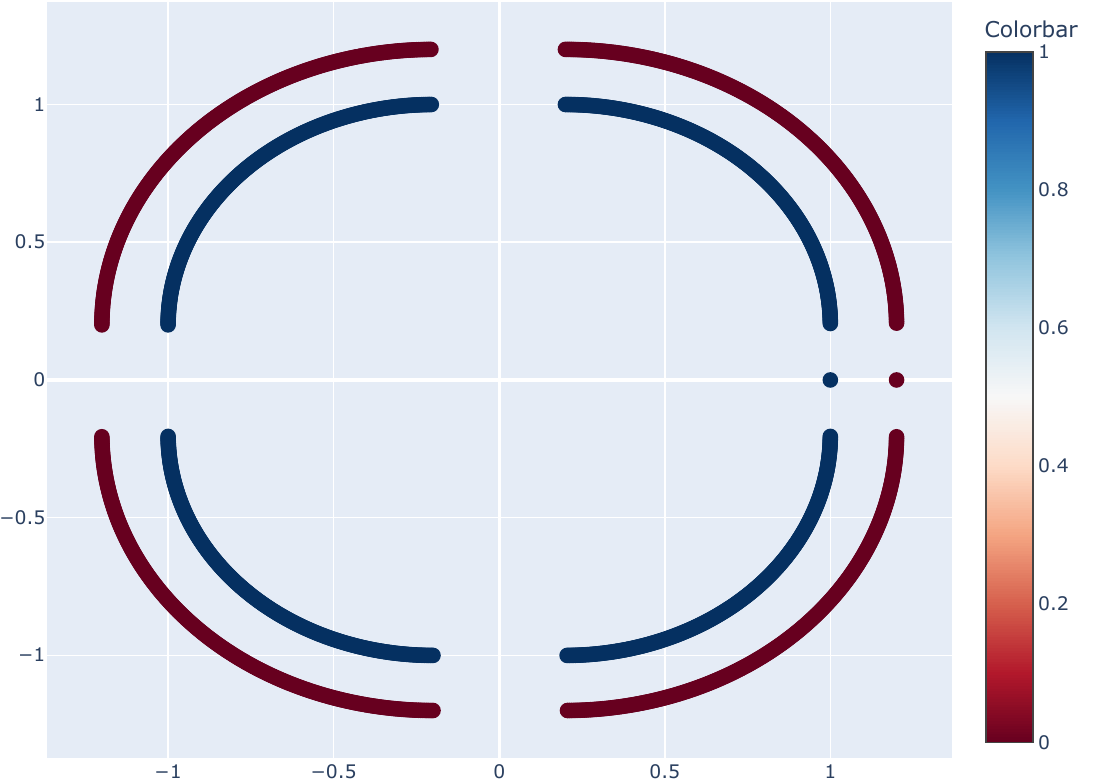}
    \caption{\(\signsplit\) with \(c = 0.2\)}
  \end{subfigure}
  \begin{subfigure}[b]{0.24\textwidth}
    \centering
    \includegraphics[width=\linewidth]{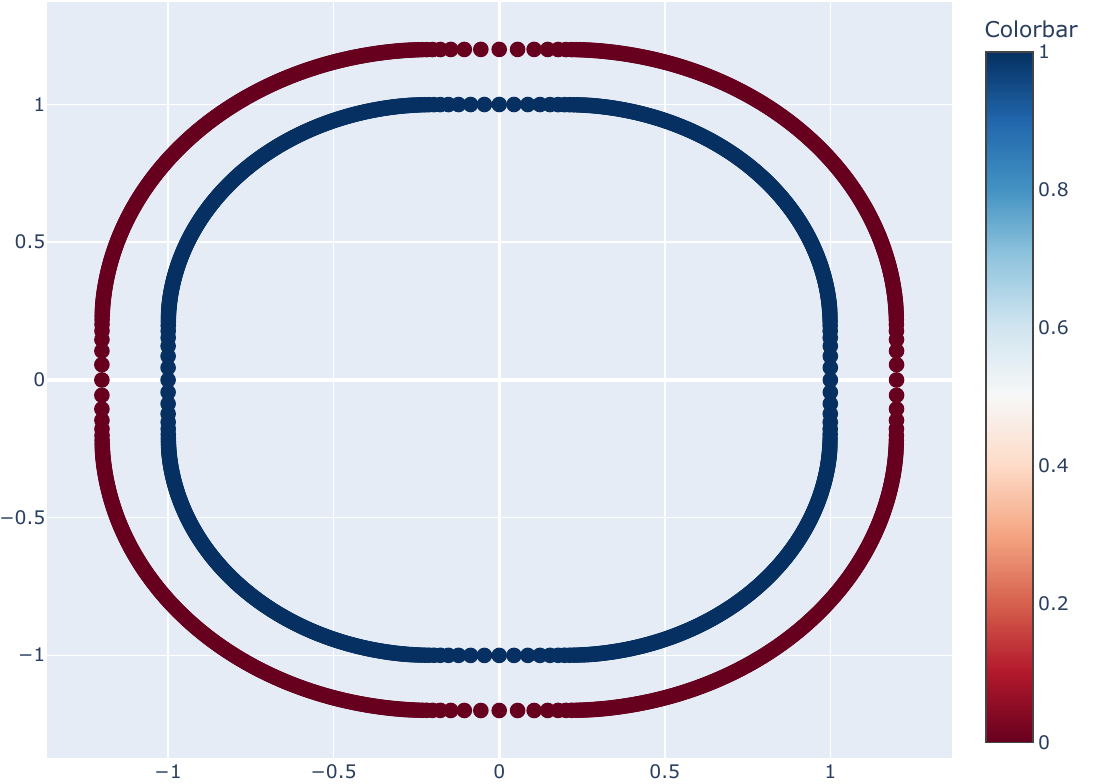}
    \caption{\(\smoothsplit\) with \(c = 0.2,\, \alpha = 40\)}
  \end{subfigure}
  \begin{subfigure}[b]{0.24\textwidth}
    \centering
    \includegraphics[width=\linewidth]{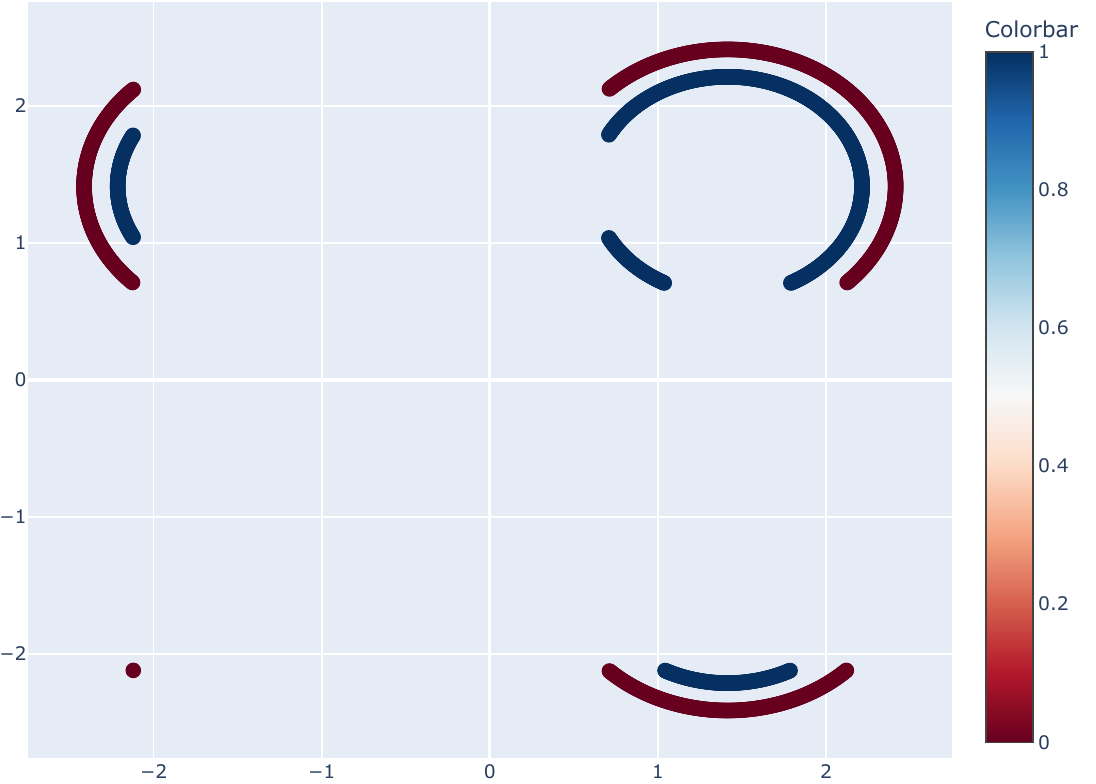}
    \caption{\(\parametricsplit\) with \(a = \frac{3\pi}{4},\, b = 1\)}
  \end{subfigure}
  \caption{Comparison of different transformations applied to the Circles dataset.}
  \label{fig:four-images}
\end{figure*}

In summary, there are two primary types of manifold deformations: compression and splitting. While \(\relu\) primarily compresses the data manifold, \(\signsplit\) and \(\smoothsplit\) effect splitting. The proposed \(\parametricsplit\) function unifies these operations by enabling both compression and splitting, depending on its parameter settings. Figure~\ref{fig:four-images} illustrates these transformations on the Circles dataset. Since the parameters of \(\parametricsplit\) are learnable during training, it integrates seamlessly into the neural network pipeline. In the following sections, we compare \(\parametricsplit\) with \(\relu\) and other widely used activation functions on both synthetic and real-world datasets.\section{Experiments}

\begin{figure}[!htb]
  \centering
  \begin{subfigure}[b]{0.45\linewidth}
    \centering
    \includegraphics[width=\linewidth]{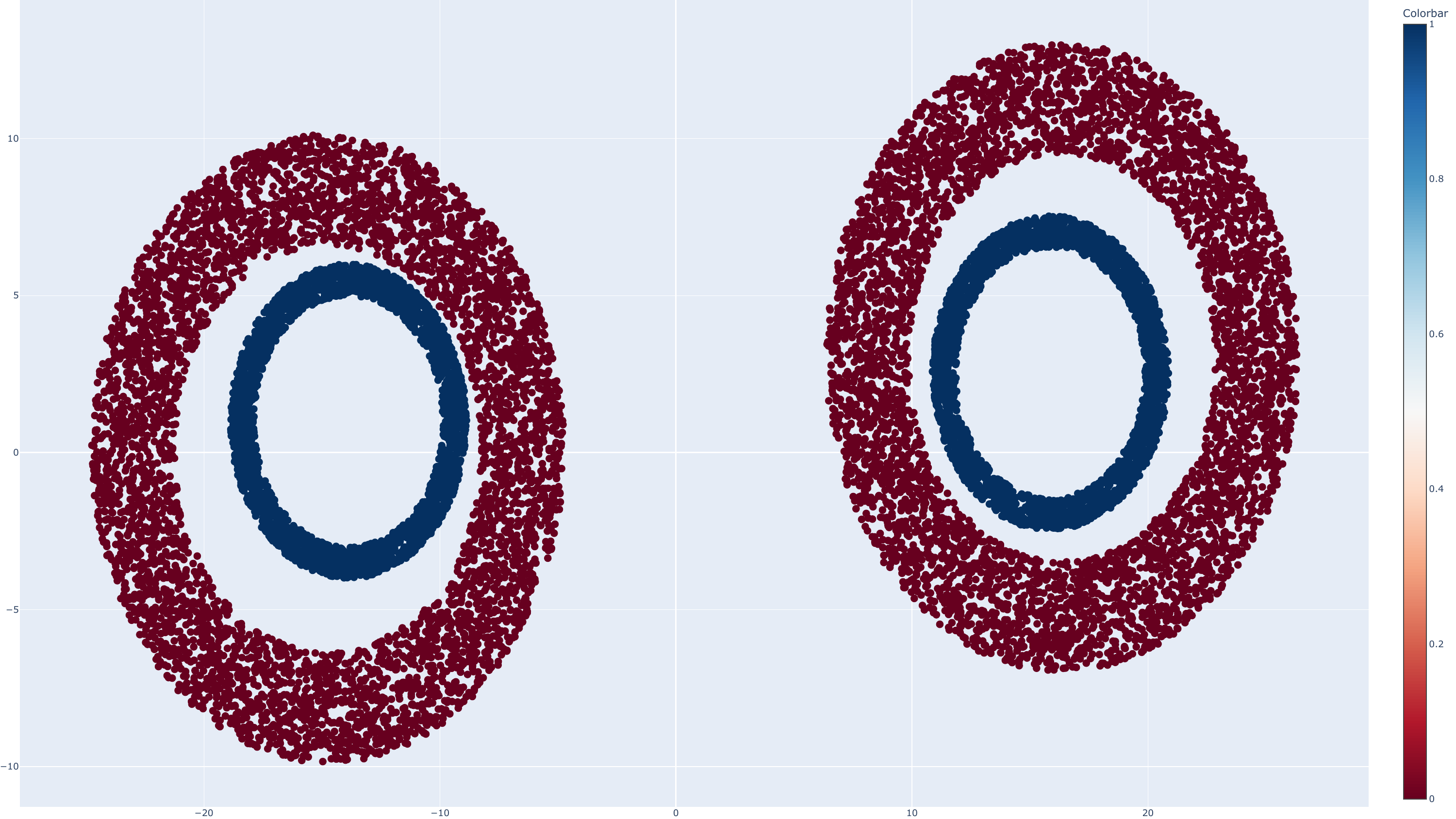}
    \caption{Dataset Circles}
  \end{subfigure}
  \begin{subfigure}[b]{0.45\linewidth}
    \centering
    \includegraphics[width=\linewidth]{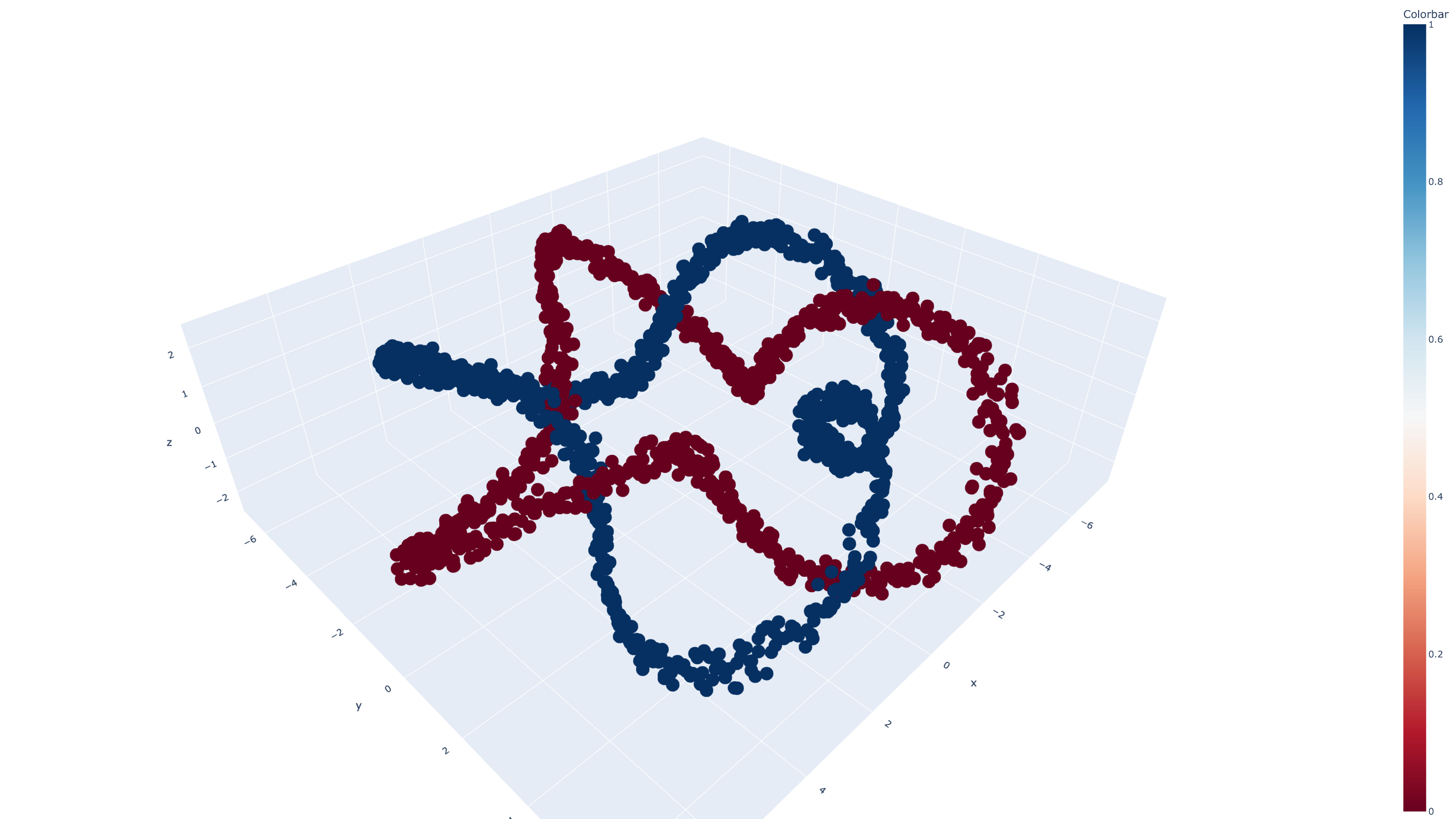}
    \caption{Dataset CurvesOnTorus}
  \end{subfigure}
  \caption{Synthetic datasets}
  \label{fig:data_manifolds}
\end{figure}

We evaluated the proposed activation functions in binary classification tasks, comparing their performance with \( \relu \), $\tanh$, and \( \prelu \). The experiments involved two synthetic datasets, CurvesOnTorus and Circles, and one real-world dataset, the Breast Cancer Wisconsin dataset. In the synthetic datasets, each class is sampled from a distinct manifold that is intertwined with others, making linear separability impossible. The data manifolds in these synthetic datasets are one-dimensional and immersed in $\mathbb{R}^2$ and $\mathbb{R}^3$ for Circles and CurvesOnTorus, respectively (illustrated in Figure~\ref{fig:data_manifolds}). As for the loss function, both in training setting and validation setting, we used binary cross-entropy loss.

We utilized fully connected neural networks for the experiments. The activation functions being compared were applied to all layers except the last one, which used \( \relu \). This design ensured that the network <<glued>> data components at the final layer, as mentioned earlier. However, further ablation studies are needed to assess the impact of this configuration.

To investigate the effects of network depth and layer dimensionality on performance, we varied the number of hidden layers and the dimensions of each layer. The number of hidden layers was set to $\{1, 2, 3\}$, and layer dimensions varied depending on the dataset: $\{2, 3, 4\}$ for Circles, $\{3, 4, 5, 6, 7\}$ for CurvesOnTorus, and $\{30, 40, 80, 100\}$ for Breast Cancer Wisconsin.

The models were trained for 100 epochs with weights initialized using Xavier normal initialization, except for networks with \( \relu \), which used Xavier uniform initialization. The learning rate was set to 0.05, and datasets were split into training and test sets in a 70/30 ratio. To ensure robustness, each experiment was repeated 10 times.

\begin{table}[!htpb]
  \centering
  \tiny
  \begin{tabularx}{\textwidth}{|>{\centering\arraybackslash}p{0.7cm}|>{\centering\arraybackslash}p{1.6cm}|*{6}{>{\centering\arraybackslash}X|}}
    \hline
    \# of layers & Activation functions & \multicolumn{3}{c|}{\textbf{Circles}}                                       & \multicolumn{3}{c|}{\textbf{CurvesOnTorus}}                                                                                                                                                                                                                                                                                                                                                         \\
                 &                      & 2                                                                           & 3                                                                           & 4                                                                           & 3                                                                           & 4                                                                           & 5                                                                           \\
    \hline
    \multirow{5}{*}{1}
                 & $\tanh$              & \makecell{$0.536$\\{$(\pm 0.066)$}}                   & \makecell{$0.506$\\{$(\pm 0.059)$}}                   & \makecell{$0.457$\\{$(\pm 0.038)$}}                   & \makecell{$0.464$\\{$(\pm 0.068)$}}                   & \makecell{$0.302$\\{$(\pm 0.056)$}}                   & \makecell{$\mathbf{0.128}$\\{$\mathbf{(\pm 0.057)}$}} \\
                 & $\relu$              & \makecell{$0.569$\\{$(\pm 0.075)$}}                   & \makecell{$0.563$\\{$(\pm 0.074)$}}                   & \makecell{$0.522$\\{$(\pm 0.053)$}}                   & \makecell{$0.599$\\{$(\pm 0.070)$}}                   & \makecell{$0.429$\\{$(\pm 0.112)$}}                   & \makecell{$0.292$\\{$(\pm 0.156)$}}                   \\
                 & $\prelu$             & \makecell{$\mathbf{0.512}$\\{$\mathbf{(\pm 0.093)}$}} & \makecell{$0.507$\\{$(\pm 0.071)$}}                   & \makecell{$0.505$\\{$(\pm 0.076)$}}                   & \makecell{$0.484$\\{$(\pm 0.112)$}}                   & \makecell{$0.337$\\{$(\pm 0.118)$}}                   & \makecell{$0.168$\\{$(\pm 0.107)$}}                   \\
                 & $\smoothsplit$       & \makecell{$0.543$\\{$(\pm 0.085)$}}                   & \makecell{$\mathbf{0.482}$\\{$\mathbf{(\pm 0.081)}$}} & \makecell{$0.461$\\{$(\pm 0.052)$}}                   & \makecell{$0.534$\\{$(\pm 0.045)$}}                   & \makecell{$0.377$\\{$(\pm 0.140)$}}                   & \makecell{$0.274$\\{$(\pm 0.106)$}}                   \\
                 & $\parametricsplit$   & \makecell{$0.527$\\{$(\pm 0.012)$}}                   & \makecell{$0.494$\\{$(\pm 0.087)$}}                   & \makecell{$\mathbf{0.398}$\\{$\mathbf{(\pm 0.128)}$}} & \makecell{$\mathbf{0.462}$\\{$\mathbf{(\pm 0.144)}$}} & \makecell{$\mathbf{0.251}$\\{$\mathbf{(\pm 0.193)}$}} & \makecell{$0.202$\\{$(\pm 0.122)$}}                   \\
    \hline
    \multirow{5}{*}{2}
                 & $\tanh$              & \makecell{$0.591$\\{$(\pm 0.048)$}}                   & \makecell{$0.530$\\{$(\pm 0.066)$}}                   & \makecell{$0.490$\\{$(\pm 0.086)$}}                   & \makecell{$0.452$\\{$(\pm 0.118)$}}                   & \makecell{$0.242$\\{$(\pm 0.091)$}}                   & \makecell{$\mathbf{0.111}$\\{$\mathbf{(\pm 0.069)}$}} \\
                 & $\relu$              & \makecell{$0.576$\\{$(\pm 0.064)$}}                   & \makecell{$0.513$\\{$(\pm 0.096)$}}                   & \makecell{$0.511$\\{$(\pm 0.123)$}}                   & \makecell{$0.504$\\{$(\pm 0.105)$}}                   & \makecell{$0.383$\\{$(\pm 0.134)$}}                   & \makecell{$0.283$\\{$(\pm 0.120)$}}                   \\
                 & $\prelu$             & \makecell{$\mathbf{0.511}$\\{$\mathbf{(\pm 0.140)}$}} & \makecell{$0.462$\\{$(\pm 0.087)$}}                   & \makecell{$0.428$\\{$(\pm 0.073)$}}                   & \makecell{$0.426$\\{$(\pm 0.090)$}}                   & \makecell{$0.230$\\{$(\pm 0.160)$}}                   & \makecell{$0.155$\\{$(\pm 0.127)$}}                   \\
                 & $\smoothsplit$       & \makecell{$0.556$\\{$(\pm 0.068)$}}                   & \makecell{$0.517$\\{$(\pm 0.084)$}}                   & \makecell{$0.496$\\{$(\pm 0.049)$}}                   & \makecell{$0.479$\\{$(\pm 0.078)$}}                   & \makecell{$0.413$\\{$(\pm 0.121)$}}                   & \makecell{$0.208$\\{$(\pm 0.122)$}}                   \\
                 & $\parametricsplit$   & \makecell{$0.555$\\{$(\pm 0.076)$}}                   & \makecell{$\mathbf{0.455}$\\{$\mathbf{(\pm 0.142)}$}} & \makecell{$\mathbf{0.388}$\\{$\mathbf{(\pm 0.152)}$}} & \makecell{$\mathbf{0.295}$\\{$\mathbf{(\pm 0.132)}$}} & \makecell{$\mathbf{0.217}$\\{$\mathbf{(\pm 0.193)}$}} & \makecell{$0.172$\\{$(\pm 0.211)$}}                   \\
    \hline
    \multirow{5}{*}{3}
                 & $\tanh$              & \makecell{$0.538$\\{$(\pm 0.084)$}}                   & \makecell{$0.501$\\{$(\pm 0.069)$}}                   & \makecell{$0.462$\\{$(\pm 0.109)$}}                   & \makecell{$0.462$\\{$(\pm 0.093)$}}                   & \makecell{$\mathbf{0.288}$\\{$\mathbf{(\pm 0.110)}$}} & \makecell{$\mathbf{0.132}$\\{$\mathbf{(\pm 0.055)}$}} \\
                 & $\relu$              & \makecell{$0.601$\\{$(\pm 0.048)$}}                   & \makecell{$0.506$\\{$(\pm 0.076)$}}                   & \makecell{$0.479$\\{$(\pm 0.116)$}}                   & \makecell{$0.521$\\{$(\pm 0.085)$}}                   & \makecell{$0.483$\\{$(\pm 0.173)$}}                   & \makecell{$0.241$\\{$(\pm 0.188)$}}                   \\
                 & $\prelu$             & \makecell{$0.530$\\{$(\pm 0.079)$}}                   & \makecell{$\mathbf{0.413}$\\{$\mathbf{(\pm 0.069)}$}} & \makecell{$0.450$\\{$(\pm 0.106)$}}                   & \makecell{$0.491$\\{$(\pm 0.097)$}}                   & \makecell{$0.330$\\{$(\pm 0.220)$}}                   & \makecell{$0.171$\\{$(\pm 0.148)$}}                   \\
                 & $\smoothsplit$       & \makecell{$0.579$\\{$(\pm 0.060)$}}                   & \makecell{$0.521$\\{$(\pm 0.068)$}}                   & \makecell{$0.485$\\{$(\pm 0.052)$}}                   & \makecell{$0.500$\\{$(\pm 0.124)$}}                   & \makecell{$0.385$\\{$(\pm 0.110)$}}                   & \makecell{$0.312$\\{$(\pm 0.114)$}}                   \\
                 & $\parametricsplit$   & \makecell{$\mathbf{0.516}$\\{$\mathbf{(\pm 0.086)}$}} & \makecell{$0.531$\\{$(\pm 0.063)$}}                   & \makecell{$\mathbf{0.385}$\\{$\mathbf{(\pm 0.171)}$}} & \makecell{$\mathbf{0.387}$\\{$\mathbf{(\pm 0.168)}$}} & \makecell{$0.326$\\{$(\pm 0.236)$}}                   & \makecell{$0.134$\\{$(\pm 0.158)$}}                   \\
    \hline
  \end{tabularx}

  \vspace{0em}  

  \begin{tabularx}{\textwidth}{|>{\centering\arraybackslash}p{0.7cm}|>{\centering\arraybackslash}p{1.6cm}|*{6}{>{\centering\arraybackslash}X|}}
    \hline
    \# of layers & Activation functions & \multicolumn{2}{c|}{\textbf{CurvesOnTorus}}                                 & \multicolumn{4}{c|}{\textbf{Breast Cancer}}                                                                                                                                                                                                                                                                                                                                                         \\
                 &                      & 6                                                                           & 7                                                                           & 30                                                                          & 40                                                                          & 80                                                                          & 100                                                                         \\
    \hline
    \multirow{5}{*}{1}
                 & $\tanh$              & \makecell{$\mathbf{0.080}$\\{$\mathbf{(\pm 0.083)}$}} & \makecell{$\mathbf{0.044}$\\{$\mathbf{(\pm 0.036)}$}} & \makecell{$0.659$\\{$(\pm 0.012)$}}                   & \makecell{$0.659$\\{$(\pm 0.013)$}}                   & \makecell{$0.659$\\{$(\pm 0.012)$}}                   & \makecell{$0.659$\\{$(\pm 0.012)$}}                   \\
                 & $\relu$              & \makecell{$0.226$\\{$(\pm 0.124)$}}                   & \makecell{$0.063$\\{$(\pm 0.050)$}}                   & \makecell{$0.458$\\{$(\pm 0.213)$}}                   & \makecell{$0.299$\\{$(\pm 0.207)$}}                   & \makecell{$0.487$\\{$(\pm 0.224)$}}                   & \makecell{$0.534$\\{$(\pm 0.205)$}}                   \\
                 & $\prelu$             & \makecell{$0.104$\\{$(\pm 0.120)$}}                   & \makecell{$0.072$\\{$(\pm 0.103)$}}                   & \makecell{$0.358$\\{$(\pm 0.214)$}}                   & \makecell{$0.364$\\{$(\pm 0.209)$}}                   & \makecell{$0.314$\\{$(\pm 0.189)$}}                   & \makecell{$\mathbf{0.286}$\\{$\mathbf{(\pm 0.137)}$}} \\
                 & $\smoothsplit$       & \makecell{$0.225$\\{$(\pm 0.091)$}}                   & \makecell{$0.122$\\{$(\pm 0.096)$}}                   & \makecell{$0.490$\\{$(\pm 0.214)$}}                   & \makecell{$0.383$\\{$(\pm 0.176)$}}                   & \makecell{$\mathbf{0.292}$\\{$\mathbf{(\pm 0.148)}$}} & \makecell{$0.416$\\{$(\pm 0.220)$}}                   \\
                 & $\parametricsplit$   & \makecell{$0.144$\\{$(\pm 0.089)$}}                   & \makecell{$0.111$\\{$(\pm 0.106)$}}                   & \makecell{$\mathbf{0.327}$\\{$\mathbf{(\pm 0.190)}$}} & \makecell{$\mathbf{0.230}$\\{$\mathbf{(\pm 0.066)}$}} & \makecell{$0.443$\\{$(\pm 0.228)$}}                   & \makecell{$0.352$\\{$(\pm 0.183)$}}                   \\
    \hline
    \multirow{5}{*}{2}
                 & $\tanh$              & \makecell{$0.076$\\{$(\pm 0.062)$}}                   & \makecell{$0.067$\\{$(\pm 0.027)$}}                   & \makecell{$0.659$\\{$(\pm 0.012)$}}                   & \makecell{$0.659$\\{$(\pm 0.013)$}}                   & \makecell{$0.659$\\{$(\pm 0.013)$}}                   & \makecell{$0.659$\\{$(\pm 0.012)$}}                   \\
                 & $\relu$              & \makecell{$0.169$\\{$(\pm 0.130)$}}                   & \makecell{$0.088$\\{$(\pm 0.101)$}}                   & \makecell{$\mathbf{0.247}$\\{$\mathbf{(\pm 0.149)}$}} & \makecell{$0.334$\\{$(\pm 0.214)$}}                   & \makecell{$\mathbf{0.333}$\\{$\mathbf{(\pm 0.227)}$}} & \makecell{$\mathbf{0.381}$\\{$\mathbf{(\pm 0.248)}$}} \\
                 & $\prelu$             & \makecell{$0.072$\\{$(\pm 0.061)$}}                   & \makecell{$\mathbf{0.014}$\\{$\mathbf{(\pm 0.014)}$}} & \makecell{$0.354$\\{$(\pm 0.219)$}}                   & \makecell{$0.315$\\{$(\pm 0.183)$}}                   & \makecell{$0.419$\\{$(\pm 0.183)$}}                   & \makecell{$0.617$\\{$(\pm 0.131)$}}                   \\
                 & $\smoothsplit$       & \makecell{$0.248$\\{$(\pm 0.129)$}}                   & \makecell{$0.097$\\{$(\pm 0.095)$}}                   & \makecell{$0.614$\\{$(\pm 1.113)$}}                   & \makecell{$0.533$\\{$(\pm 0.339)$}}                   & \makecell{$0.450$\\{$(\pm 0.227)$}}                   & \makecell{$2.971$\\{$(\pm 5.311)$}}                   \\
                 & $\parametricsplit$   & \makecell{$\mathbf{0.058}$\\{$\mathbf{(\pm 0.111)}$}} & \makecell{$0.035$\\{$(\pm 0.067)$}}                   & \makecell{$0.310$\\{$(\pm 0.192)$}}                   & \makecell{$\mathbf{0.243}$\\{$\mathbf{(\pm 0.096)}$}} & \makecell{$0.567$\\{$(\pm 0.391)$}}                   & \makecell{$0.541$\\{$(\pm 0.194)$}}                   \\
    \hline
    \multirow{5}{*}{3}
                 & $\tanh$              & \makecell{$\mathbf{0.099}$\\{$\mathbf{(\pm 0.093)}$}} & \makecell{$0.035$\\{$(\pm 0.032)$}}                   & \makecell{$0.659$\\{$(\pm 0.013)$}}                   & \makecell{$0.659$\\{$(\pm 0.012)$}}                   & \makecell{$0.659$\\{$(\pm 0.013)$}}                   & \makecell{$0.659$\\{$(\pm 0.013)$}}                   \\
                 & $\relu$              & \makecell{$0.285$\\{$(\pm 0.220)$}}                   & \makecell{$\mathbf{0.032}$\\{$\mathbf{(\pm 0.034)}$}} & \makecell{$0.341$\\{$(\pm 0.229)$}}                   & \makecell{$0.393$\\{$(\pm 0.229)$}}                   & \makecell{$\mathbf{0.387}$\\{$\mathbf{(\pm 0.230)}$}} & \makecell{$0.506$\\{$(\pm 0.205)$}}                   \\
                 & $\prelu$             & \makecell{$0.307$\\{$(\pm 0.462)$}}                   & \makecell{$0.089$\\{$(\pm 0.072)$}}                   & \makecell{$0.310$\\{$(\pm 0.189)$}}                   & \makecell{$\mathbf{0.376}$\\{$\mathbf{(\pm 0.206)}$}} & \makecell{$0.617$\\{$(\pm 0.164)$}}                   & \makecell{$0.527$\\{$(\pm 0.223)$}}                   \\
                 & $\smoothsplit$       & \makecell{$0.300$\\{$(\pm 0.245)$}}                   & \makecell{$0.268$\\{$(\pm 0.109)$}}                   & \makecell{$0.574$\\{$(\pm 0.370)$}}                   & \makecell{$0.596$\\{$(\pm 0.545)$}}                   & \makecell{$8.133$\\{$(\pm 18.376)$}}                  & \makecell{$6.083$\\{$(\pm 14.711)$}}                  \\
                 & $\parametricsplit$   & \makecell{$0.127$\\{$(\pm 0.190)$}}                   & \makecell{$0.085$\\{$(\pm 0.177)$}}                   & \makecell{$\mathbf{0.283}$\\{$\mathbf{(\pm 0.161)}$}} & \makecell{$0.386$\\{$(\pm 0.216)$}}                   & \makecell{$0.465$\\{$(\pm 0.272)$}}                   & \makecell{$\mathbf{0.500}$\\{$\mathbf{(\pm 0.255)}$}} \\
    \hline
  \end{tabularx}
  \caption{Val. loss averaged in 10 runs}
  \label{table:results}
\end{table}

\section{Results}
The experimental results summarized in Table~\ref{table:results} demonstrate that the proposed \(\parametricsplit\) activation function performs exceptionally well, particularly in low-dimensional settings. It outperforms traditional activation functions such as \(\relu\), \(\tanh\), and \(\prelu\) when the layer dimensions are small relative to the data manifold's intrinsic dimensionality. In higher-dimensional scenarios, \(\parametricsplit\) achieves performance comparable to the best-performing conventional activations. Similarly, \(\smoothsplit\) delivers competitive results, often matching those of \(\relu\) and \(\tanh\).

A notable observation is the strong performance of \(\tanh\) and \(\prelu\) on the CurvesOnTorus dataset in configurations with larger dimensions. This phenomenon is likely due to the one-dimensional nature of the class manifolds in this dataset; immersion in \(\mathbb{R}^5\) or higher affords the network sufficient degrees of freedom to disentangle the curves via homeomorphic transformations. Conversely, in low-dimensional settings, the network benefits from explicitly manipulating the data topology. The learnable parameters of \(\parametricsplit\) provide the necessary flexibility to <<cut>> the topology and adapt to the constraints imposed by limited layer dimensions.

\section{Conclusion}
In this study, we introduced novel activation functions, \(\parametricsplit\) and \(\smoothsplit\), which enhance neural networks by enabling explicit manipulation of data topology. These functions extend the capabilities of traditional activations such as \(\relu\) by facilitating both topological <<cutting>> and <<gluing>> operations, thereby offering greater flexibility in adapting to the underlying data manifold. Our experiments on synthetic and real-world datasets demonstrate that \(\parametricsplit\) consistently outperforms conventional activations in low-dimensional settings—where topological transformations are critical—while remaining competitive in higher dimensions.

Future work will explore the broader applicability of these functions in diverse machine learning tasks and assess the impact of topological transformations on network generalization. Additionally, integrating these activation functions into more complex architectures, such as convolutional or transformer-based models, represents an interesting direction for further research.

\begin{footnotesize}
  \bibliographystyle{unsrt}
  \bibliography{references}
\end{footnotesize}
\end{document}